\documentclass[sigconf]{acmart}
\AtBeginDocument{%
  \providecommand\BibTeX{{%
    \normalfont B\kern-0.5em{\scshape i\kern-0.25em b}\kern-0.8em\TeX}}}

\setcopyright{acmcopyright}
\copyrightyear{2022}
\acmYear{2022}
\acmDOI{XXXXXXX.XXXXXXX}

\acmConference[Conference acronym 'XX]{Make sure to enter the correct
  conference title from your rights confirmation emai}{June 03--05,
  2018}{Woodstock, NY}
\acmPrice{15.00}
\acmISBN{978-1-4503-XXXX-X/18/06}

\usepackage{microtype}
\usepackage{graphicx}
 
\usepackage{amsmath,amssymb,amsfonts}
\usepackage{subfigure}

\usepackage{booktabs} % for professional tables
\usepackage{adjustbox}
\usepackage{caption}

\usepackage{hyperref}
\newtheorem{theorem}{Theorem}

\usepackage{bm}
\usepackage{multirow}

\usepackage{cases}
\usepackage{enumitem}

\usepackage{xcolor}

\begin{document}

\title{Test-Time Training for Graph Neural Networks}

% \affiliation{%

% \affiliation{%

% \affiliation{%



\author{Yiqi Wang}
\affiliation{%
 \institution{Michigan State University}
 \country{the United States}
}
 \email{wangy206@msu.edu}

\author{Chaozhuo Li}
\affiliation{%
 \institution{Microsoft Research Asia}
 \country{China}
}
\email{cli@microsoft.com}

\author{Wei Jin}
\affiliation{%
 \institution{Michigan State University}
 \country{the United States}
}
 \email{jinwei2@msu.edu}

\author{Rui Li}
\affiliation{%
 \institution{Dalian University of Technology }
 \country{China}
}
 \email{lirui121200@mail.dlut.edu.cn}

\author{Jianan Zhao}
\affiliation{%
 \institution{University of Notre Dame}
 \country{the United States}
}
 \email{andy.zhaoja@gmail.com}

\author{Jiliang Tang}
\affiliation{%
 \institution{Michigan State University}
  \country{the United States}
}
 \email{tangjili@msu.edu}
 
\author{Xing Xie}
\affiliation{%
 \institution{Microsoft Research Asia}
 \country{China}
}
\email{xingx@microsoft.com}

\begin{abstract}
Graph Neural Networks (GNNs) have made tremendous progress in the graph classification task. However, a performance gap between the training set and the test set has often been noticed. To bridge such gap, in this work we introduce the first test-time training framework for GNNs to enhance the model generalization capacity for the graph classification task. In particular, we design a novel test-time training strategy with self-supervised learning to adjust the GNN model for each test graph sample. Experiments on the benchmark datasets have  demonstrated the effectiveness of the proposed framework, especially when there are distribution shifts between training set and test set. We have also conducted exploratory studies and theoretical analysis to gain deeper understandings on the rationality of  the design of the proposed graph test time training framework (GT3). 
\end{abstract}





\maketitle

\section{Introduction}

Many real-world data can be naturally represented as graphs, such as social networks~\cite{yanardag2015deep,fan2019graph}, molecules~\cite{borgwardt2005protein,ying2018hierarchical,hu2020open} and single-cell data~\cite{wang2021scgnn,wen2022graph}. Graph neural networks (GNNs)~\cite{wu2019comprehensive-survey,ma2021deep,battaglia2018relational,liu2021elastic}, a successful generalization of deep neural networks (DNNs) over the graph domain, typically learn graph representations by fusing the node attributes and topological structures. 
GNNs have been empirically and theoretically proven to be effective in graph representation learning, and have achieved revolutionary progress in various graph-related tasks, including node classification~\cite{kipf2016semi,liu2021elastic,gat}, graph classification~\cite{ying2018hierarchical,xu2018how,ma2019graph} and link prediction~\cite{vgae,wang2022localized}. 
\begin{table}[]

\centering
\caption{Graph classification results on \textit{ogbg-molhiv}. The learned GNN model is applied on training, validation, and test sets, respectively. (Results are from the paper~\cite{hu2020open})}
\begin{adjustbox}{width = 0.45\textwidth}

  \setlength{\tabcolsep}{4mm}{
\begin{tabular}{cccc}

\hline
\multirow{2}{*}{\textbf{Method}} & \multicolumn{3}{c}{\textbf{ROC-AUC}(\%)} \\ 
     \cmidrule(l){2-4} & Training & Validation  & Test    \\ \hline

GCN  & 88.65\scriptsize$\pm{1.01}$    & 83.73\scriptsize$\pm{0.78}$          & 74.18\scriptsize$\pm{1.22}$   \\ \hline
GIN   &93.89\scriptsize$\pm{2.96}$         &84.1\scriptsize$\pm{1.05}$          &75.2\scriptsize$\pm{1.30}$     \\ \hline

\end{tabular}
}
 \end{adjustbox}
\label{tab:ogbg-molhiv}

\end{table}

\begin{figure}
  \centering
  \includegraphics[scale=0.25]{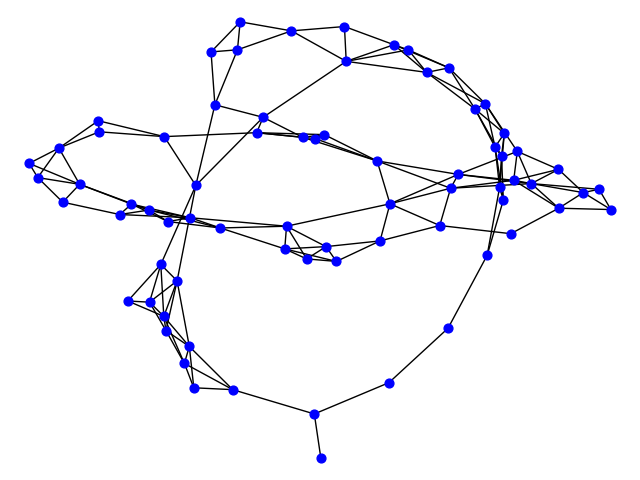}
  \includegraphics[scale=0.25]{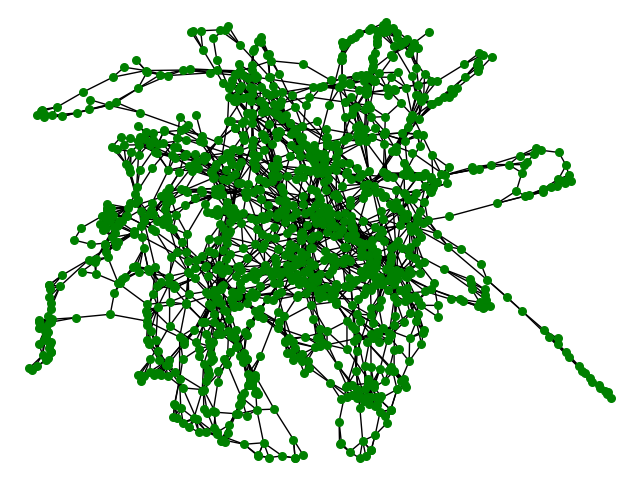}
  \caption{Two graph samples from D\&D dataset~\cite{morris2020tudataset}, which present very different structure properties}
  \label{fig:DD-intro}
\end{figure}
As one of the most crucial tasks on graphs, graph classification aims at identifying the class labels of graphs based on the graph properties. 
Graph classification has been widely employed in a myriad of real-world scenarios, such as drug discovery~\cite{duvenaud2015convolutional}, protein prediction~\cite{gligorijevic2021structure} and single-cell analysis~\cite{wang2021scgnn,wen2022graph}. 
Existing graph classification models generally follow the training-test paradigm. 
Namely, a graph neural network is first trained on the training set to learn the correlations between annotations and the latent topological structures (e.g., Motif~\cite{zhang2021motif} and Graphlets~\cite{bouritsas2022improving}), and then apply the captured knowledge on the test samples to make classifications. 
The nucleus of such graph classification paradigm is the compatibility between training and test set, in which case the learned knowledge from the training set is capable of smoothly providing accurate predicting signals for test samples.     
However, this assumption may not hold in many real scenarios~\cite{sun2019test,nagarajan2020understanding}, especially on graph tasks~\cite{hu2020open}. 
As shown in Table~\ref{tab:ogbg-molhiv}, there exist significant gaps between the performance of two popular GNN models (GCN~\cite{kipf2016semi} and GIN~\cite{xu2018how}) in the training and test set. 
The underlying reason lies in that graphs are non-Euclidean data structures defined in  high-dimensional space, leading to infinite and diverse possible topological structures. 
For example, as demonstrated in Figure~\ref{fig:DD-intro}, two graph samples with the same labels from the same dataset present totally different structural patterns. 
Apparently, it is intractable to directly transfer the knowledge from the left graph to the right one. 

In this paper, we argue that the learned knowledge within the trained GNN models should be delicately calibrated to adapt to the unique characteristics of each test sample. 
Inspired by recent progress in computer vision~\cite{sun2019test,liu2021ttt++}, we further go beyond the traditional training-test paradigm to the three-step mechanism: training, test-time training, and test, for the graph neural networks.  
The test-time training is designed to incorporate the unique features from each test graph sample to adjust the trained GNN model. 
The essence of test-time training is to utilize the auxiliary unsupervised self-supervised learning (SSL) tasks to give a hint about the underlying traits of a test sample and then leverage this hint to calibrate the trained GNN model for this test sample. 
Given the distinct properties in each graph sample, test-time training has great potential to improve the generalization of GNN models. 

However, it is challenging to design a test-time training framework for graph neural networks over graph data. First, how to design effective self-supervised learning tasks specialized for graph classification is obscure. 
Graph data usually consists of not only node attribute information but also abundant structure information. Thus, it is important to blend these two kinds of graph information into the design of the self-supervised learning task. 
Second, how to mitigate the potential severe feature space distortion during the test time? The representation space can be severally tortured in some directions that are undesirable for the main task in the test-time training~\cite{liu2021ttt++}, since the model is adapted solely based on a single sample. 
This challenge of distortion is more severe for graphs considering the diverse and volatile topological structures. 
Last but not least, it is desired to throw light on the rationality of test-time training over GNNs.

In order to solve these challenges, we propose Graph Test-Time Training with Constraint (GT3), a test-time training framework to improve the performance of GNN models on graph classification tasks. 
Specifically, a hierarchical self-supervised learning framework is specifically defined for test-time training on graphs. 
To avoid the risk of distribution distortion, we propose a new adaptation constraint to seek a proper equilibrium between the knowledge from the training set and the unique signals from the single test sample. 
Furthermore, extensive exploratory studies and theoretical analysis are conducted to understand the rationality of the GT3 model. 
Our proposal is extensively evaluated over a set of datasets with various basis models, demonstrating the superiority of the GT3 model.  

Our key contributions are summarized as follows:
\begin{itemize}
    \item We investigate the novel task of test-time training on GNNs for graph classification, which is capable of simultaneously enjoying the merits of the training knowledge and the characteristics of the test sample.    

    \item We propose a novel model GT3 for test-time training in GNNs, in which various SSL tasks and adaption constraints are elaborately designed to boost classification performance and GNN generalization ability.  
    
    \item  We conduct extensive experiments over four datasets and our proposal consistently achieves SOTA performance over all the datasets.

    \item We have  conducted exploratory studies and theoretical analysis to understand the rationality of the design of the proposed framework. 
\end{itemize}

\section{Preliminary}

In this section, we will introduce the key notations used in this paper by defining the graph classification problem and reviewing the general paradigm of Graph Neural Networks (GNNs).

\subsection{Graph Classification}
Graph classification is one of the most common graph-level tasks. It aims at predicting the property categories of graphs.  Suppose that there is a training set $\mathcal{G} = \{G_1,G_2,\ldots,G_n \}$, consisting of $n$ graphs. For each graph $G_i\in\mathcal{G}$, its structure information, node attributes and label information are denoted as $G_i = ({\bf A}_i,{\bf X}_i, y_i)$, where ${\bf A}_i$ is the graph adjacency matrix, ${\bf X}_i$ represents the node attributes and $y_i$ is the label. The goal of graph classification is to learn a classification model $f=(\cdot;\bm{\theta})$ based on $\mathcal{G}$ which can be used to predict the label of a new graph $G^\prime$.

\subsection{Graph Neural Networks}

Graph neural networks (GNNs) have successfully extended deep neural networks to graph domains.  Generally, graph neural networks consist of several layers, and typically each GNN layer refines node representation by feature transforming, propagating and aggregating node representation across the graph. According to~\cite{gilmer2017neural}, the node representation update process of node $v$ in the $l$-th GNN layer of an $L$-layer GNN can be formulated as follows:
\begin{align}
    {\bf a}_v^l &= \mathit{AGGREGATE}^l({{\bf h}_u^{(l-1)},u\in \mathcal{N}_v}), l\in[L] \label{eq:gnn-agg} \\
    {\bf h}_v^l &= \mathit{COMBINE}^l({\bf h}_v^{(l-1)},{\bf a}_v^l), l\in[L] \label{eq:gnn-comb} ,
\end{align}
where $\mathit{AGGREGATE}$ is the aggregation function aggregating information from neighborhood nodes $u\in \mathcal{N}_v$ , and $\mathit{COMBINE}$ is the combination function updating the node representation in the $l$-th layer based on the aggregated information $a_v^l$ and the node representation of $v$ from the previous layer $(l-1)$. For simplicity, we summarize the node representation update process in the $l$-th GNN layer, consisting of Eqs~(\ref{eq:gnn-agg}) and~(\ref{eq:gnn-comb}), as follows:
\begin{align}
    {\bf H}^l = \mathit{GNN}({\bf A},{\bf H}^{(l-1)};{\bm \theta}_l), l\in[L] \label{eq:gnn-layer-l} ,
\end{align}
where ${\bf A}$ is the adjacency matrix, ${\bf H}^l$ denotes representations of all the nodes in the $l$-th GNN layer, and ${\bm \theta}_l$ indicates the learnable parameters in the $l$-th layer. ${\bf H}^0$ is the input node attributes ${\bf X}$.

For graph classification, an overall graph representation is summarized from the $L$-th GNN layer based on the node representations by a readout function, which will be utilized by the prediction layer to predict the label of the graph:
\begin{align}
    {\bf h}_G = \mathit{READOUT}({\bf H}^{L})\label{eq:gnn-readout} ,
\end{align}

\section{Methodology}
In this section,  we introduce the proposed framework of Graph Test-Time Training with Constraint (GT3) and detail its key components. We first present the overall framework, then describe the hierarchical SSL task and the adaptation constraint to mitigate the potential risk of feature distortion in GT3. 

\subsection{Framework}

Suppose that the graph neural network for the graph classification task consists of $L$ GNN layers, and the learnable parameters in the $l$-th layer are denoted as ${\bm \theta}_l$. 

The overall model parameters can be denoted as ${\bm \theta}_{main} = ({\bm \theta}_1,{\bm \theta}_2,\ldots,{\bm \theta}_L, {\bm \delta}_m) $, where ${\bm \delta}_m$ is the parameters in the prediction layer. In this work, we mainly focus on the graph classification task, which is named as {\it the main task} in the test-time training framework. Given the training graph set $\mathcal{G} = (G_1, G_2,\ldots,G_n)$ and its corresponding graph label set $\mathcal{Y} = (y_1,y_2,\ldots,y_n)$. The main task is to solve the following empirical risk minimization problem:
\begin{align}
    \min\limits_{{\bm \theta}_{main}}\frac{1}{n}\sum\limits_{i=1}^n\mathcal{L}_m({\bf A}_i,{\bf X}_i,y_i;{\bm \theta}_{main}), G_i\in\mathcal{G}, \label{eq:main_task} 
\end{align}
where $\mathcal{L}_m(\cdot)$ denotes the loss function for the main task.

In addition to the main task, self-supervised learning (SSL) tasks also play a vital role in the test-time training framework. As informed by~\cite{liu2021ttt++}, the SSL tasks should capture the informative graph signals. 
In addition, a desirable SSL task also should be coherent to the main task, ensuring the unsupervised test-time training contributes to improving classifications.   
Here we design a delicate hierarchical constraining learning to capture the underlying node-node and node-graph correlations (details in Section~\ref{model:auxi}). 
For simplicity, the objective of the SSL task is denoted as  $\mathcal{L}_s(\cdot)$.
It consists of three major phases: (1) the training phase (Figure~\ref{fig:ttt-frame-training}); (2) the test-time training phase (Figure~\ref{fig:ttt-frame-ttt}) and (3) the test phase (Figure~\ref{fig:ttt-frame-infer}).  

\subsubsection{The Training Phase.} 
\begin{figure}[ht]%

    \centering

    \subfigure[The Training Phase. \label{fig:ttt-frame-training}]{{\includegraphics[width=1.0\linewidth, trim = 0cm 0cm 0cm 2.2cm, clip]{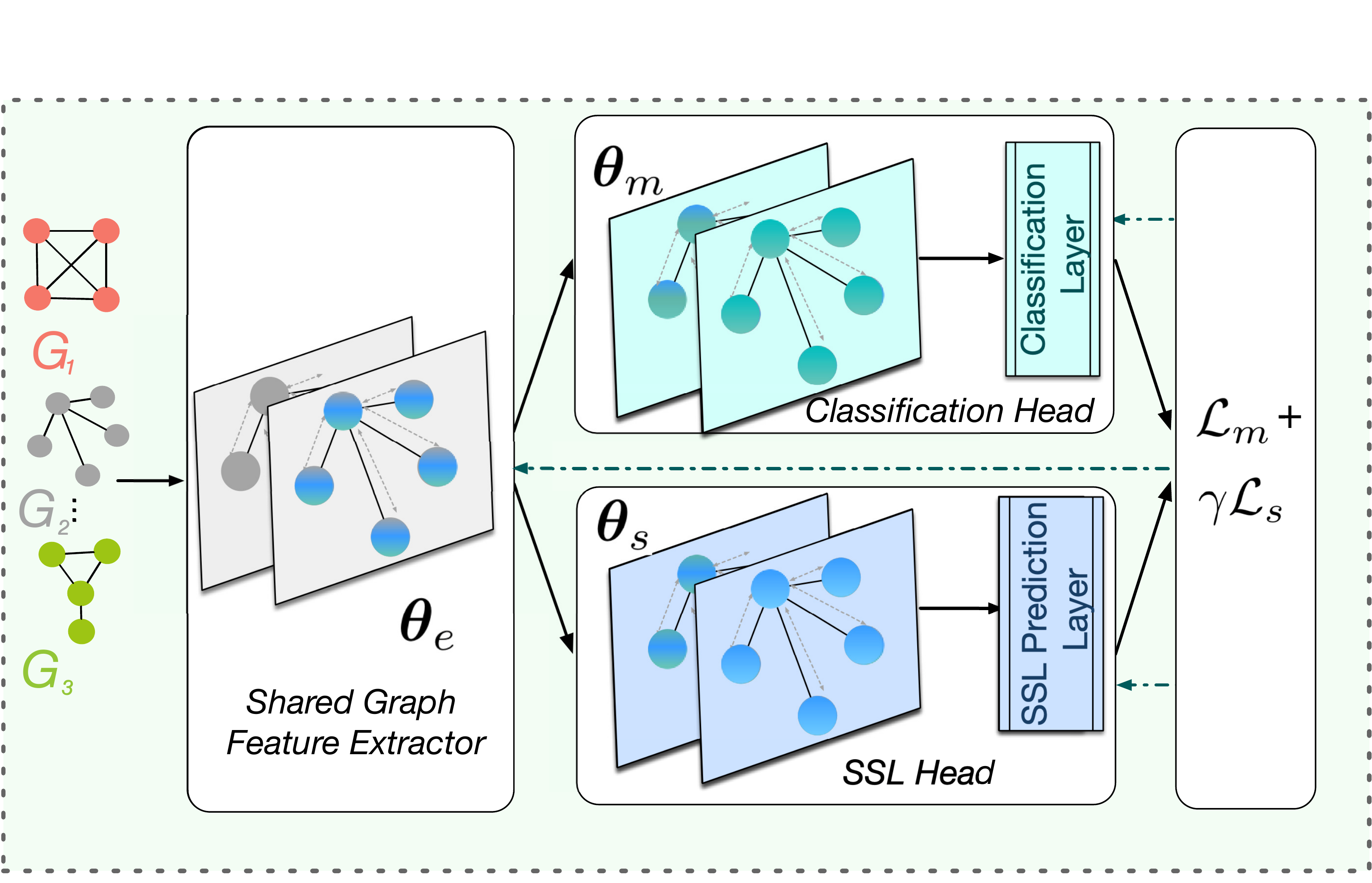}}}%
    
    \subfigure[Test-time Training Phase. \label{fig:ttt-frame-ttt}]{{\includegraphics[width=1.0\linewidth, trim = 0cm 0cm 0cm 1.5cm, clip]{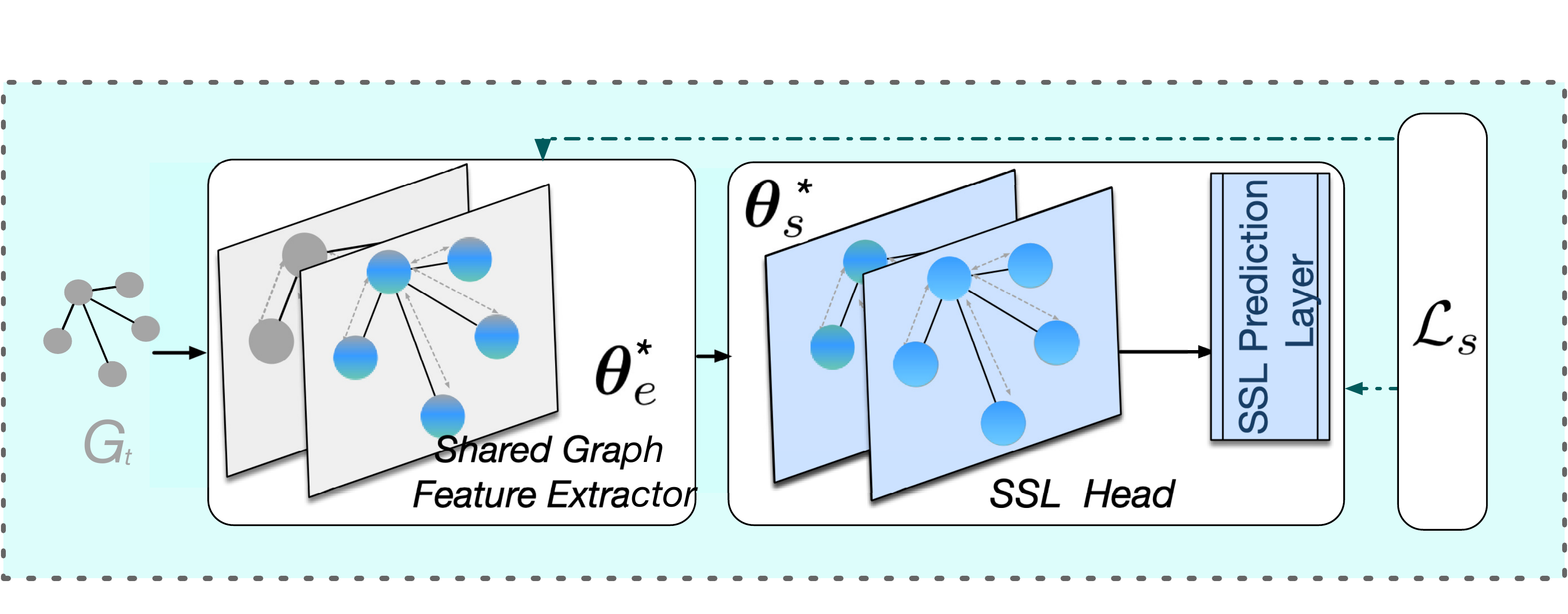} }}%
    
   \subfigure[Test Phase. \label{fig:ttt-frame-infer}]{{\includegraphics[width=1.0\linewidth, trim = 0cm 0cm 0cm 1.0cm, clip]{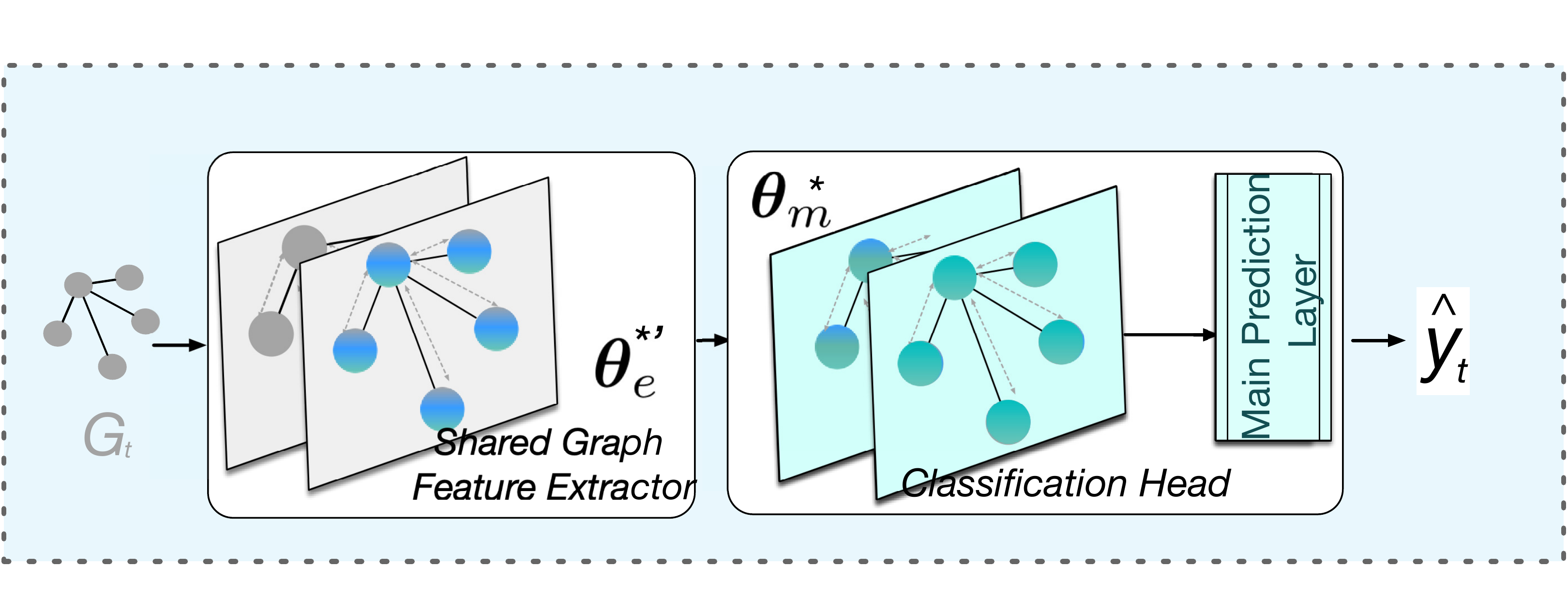} }}%

    \caption{An Overview of the Proposed Framework GT3.}%
    \label{fig:ttt-gnn-frame}

\end{figure}

Given the training set $\mathcal{G}$, the training phase is to train the GNN parameters based on the main task and the SSL task. 
The key challenge is how to reasonably integrate the main task and the SSL task for the GNN models in the test-time training framework. To tackle this challenge, we conduct an empirical study where we train two GNN models separately by the main task and the SSL task, and we found that these two GNN models will extract similar features by their first few layers. More details can be found in Section~\ref{exp:layer-fea-explore}. This empirical finding paves us a way to integrate the main task and the SSL task: we allow them to share the parameters in the first few layers as shown in Figure~\ref{fig:ttt-frame-training}. We denote the shared model parameters as ${\bm \theta}_e = ({\bm \theta}_1,{\bm \theta}_2,\ldots,{\bm \theta}_K)$, where $K \in \{1,2,\ldots,L\}$. The shared component is named as \textit{shared graph feature extractor} in this work. The model parameters that are specially for the main task are denoted as ${\bm \theta}_m = ({\bm \theta}_{K+1},\ldots,{\bm \theta}_L, \delta_m)$, described as \textit{the graph classification head}. Correspondingly, the SSL task has its own \textit{self-supervised learning head}, denoted as ${\bm \theta}_s = ({\bm \theta}_{K+1}^{'},\ldots,{\bm \theta}_L^{'}, {\bm \delta}_s)$, where ${\bm \delta}_s$ is the prediction layer for the SSL task. To summarize, the model parameters for the main task can be denoted as ${\bm \theta}_{main} = ({\bm \theta}_e,{\bm \theta}_m)$,  the model parameters for the SSL task can be represented as ${\bm \theta}_{self} = ({\bm \theta}_e,{\bm \theta}_s)$, and these for the whole framework can be denoted as ${\bm \theta}_{overall} = ({\bm \theta}_e,{\bm \theta}_m,{\bm \theta}_s)$. In the training process, we train the whole framework by minimizing a weighted combination loss of $\mathcal{L}_m(\cdot)$ and $\mathcal{L}_s(\cdot)$ as follows:

\begin{align}
    \min\limits_{{\bm \theta}_e,{\bm \theta}_m,{\bm \theta}_s}\frac{1}{n}\sum\limits_{i=1}^N\mathcal{L}_m({\bf A}_i,{\bf X}_i,y_i;{\bm \theta}_e,{\bm \theta}_m)+ \nonumber \\ 
    \gamma \mathcal{L}_s({\bf A}_i,{\bf X}_i;{\bm \theta}_e,{\bm \theta}_s), G_i\in\mathcal{G},     \label{eq:training_loss}  
\end{align}

\noindent where $\gamma$ is a hyper-parameter balancing the main task and the self-supervised task in the training process.

\begin{figure*}[!t]
\begin{center}
{\includegraphics[width=0.75\linewidth]{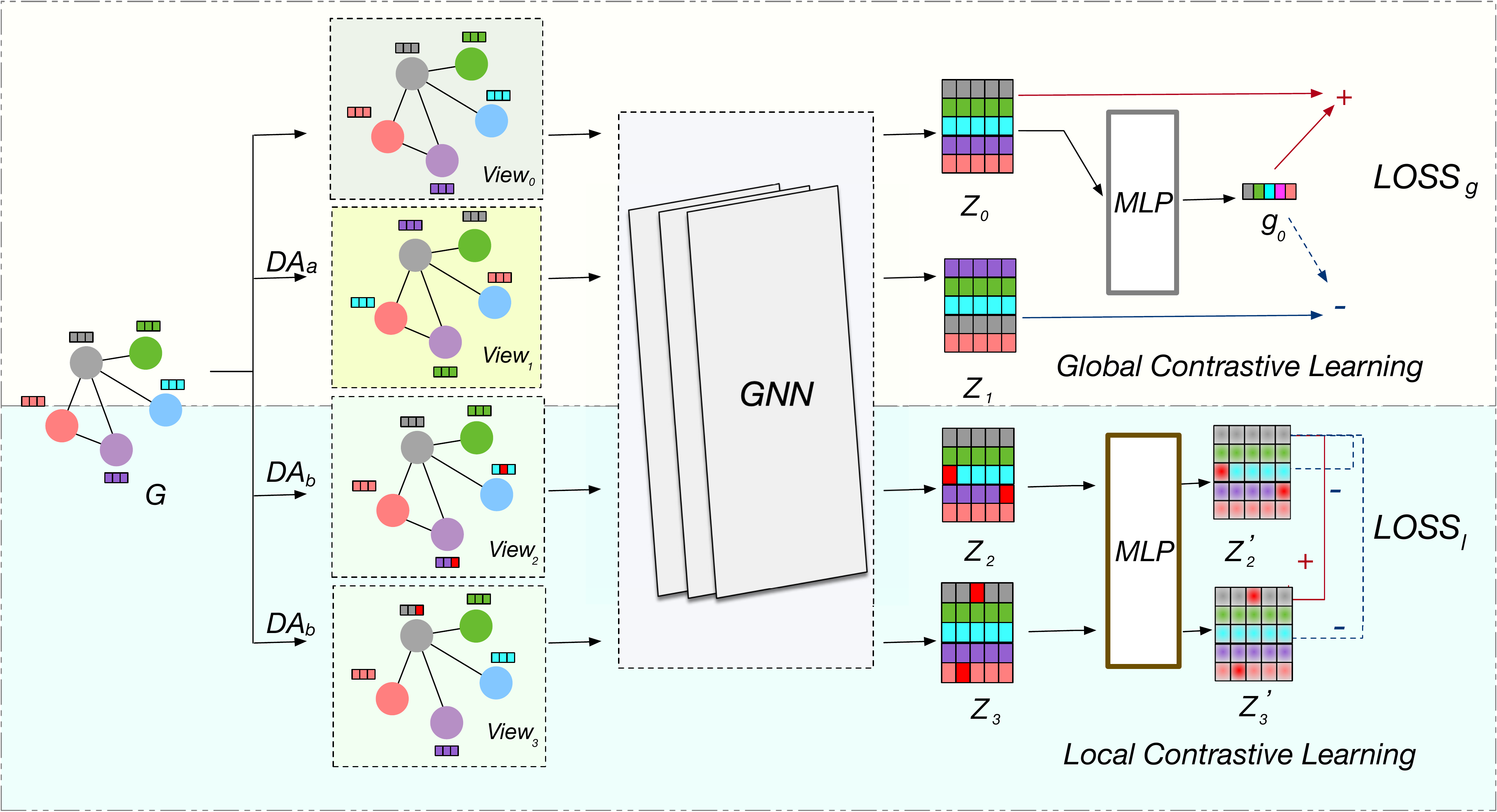}}
\end{center}

\caption{The Self-supervised Learning Task for GT3.}

\label{fig:model-frame}

\end{figure*}

\subsubsection{The Test-time Training Phase.} Given a test graph $G_t$, the test-time training process will fine-tune the learned model via the SSL task as illustrated in Figure~\ref{fig:ttt-frame-ttt}. Specifically, given a test graph sample $G_t = ({\bf A}_t, {\bf X}_t)$, the \textit{shared graph feature extractor} and the \textit{self-supervised learning head} are fine-tuned by minimizing the SSL task loss as follows:
\begin{align}
    \min\limits_{{\bm \theta}_e,{\bm \theta}_s}  \mathcal{L}_s({\bf A}_t,{\bf X}_t;{\bm \theta}_e,{\bm \theta}_s), \label{eq:test_loss} 
\end{align}
Suppose that $\bm{\theta}_{overall}^{*} = ({\bm \theta}_e^{*},{\bm \theta}_m^{*},{\bm \theta}_s^{*})$ is the optimal parameters from minimizing the overall loss of Eq~(\ref{eq:training_loss}) from the training process. 
In the test-time training phase, given a specific test graph sample $G_t$, the proposed framework further updates ${\bm \theta}_e^{*}$ and ${\bm \theta}_s^{*}$ to ${\bm \theta}_e^{*'}$ and $\bm{\theta}_s^{*'}$ by minimizing the SSL loss of Eq~(\ref{eq:test_loss}) over the test graph sample $G_t$.

\subsubsection{The Test Phase.}  The test phase is to predict the label of the test graph $G_t$ based on the fine-tuned model by the test-time training as demonstrated in Figure~\ref{fig:ttt-frame-infer}. In particular, the label of $G_t$ is predicted via the GNN model $f(\cdot;{\bm \theta}_e^{*'},{\bm \theta}_m^{*})$ as:
\begin{align}
    \hat y_t = f({\bf X}_t,{\bf A}_t;{\bm \theta}_e^{*'},{\bm \theta}_m^{*}). \label{eq:prediction} 
\end{align}

\subsection{Self-supervised Learning Task for GT3}\label{model:auxi}
An appropriate and informative SSL task is the key to the success of test-time training~\cite{liu2021ttt++}. However, it is challenging to design an appropriate SSL task for test-time training for GNNs. First, graph data is intrinsically different from images, some common properties such as rotation invariance~\cite{shorten2019survey} do not exist in graph data, thus most SSL tasks commonly adopted by existing test-time training are invalid in graph data. Second, most existing SSL tasks for graph classification are intra-graph contrastive learning based on the difference among diverse graph samples~\cite{sun2019infograph,hassani2020contrastive}. These are not applicable for the test-time training where we aim at specifically adapting the model for every single graph during the test time. Third, graph data usually consists of not only node attribute information but abundant and unique structure information. It is important to fully leverage these two kinds of graph information into the design of the SSL task since an informative SSL task is one of the keys to the success of test-time training.  

 Motivated by the success of contrastive learning in graph domains, we propose to use a hierarchical SSL task from both local and global perspectives for GT3, which fully exploits the graph  information from both the node-node level and node-graph level. 

 In addition, the proposed SSL task is not built up based on the differences among distinct graphs, thus it can be naturally applied to a single graph. We empirically validate that the global (node-graph level) contrastive learning task and the local (node-node level) contrastive learning task in the proposed two-level SSL task is necessary for GT3 since they are complementary to each other across datasets. More details about this empirical study can be found in Section~\ref{exp:layer-fea-explore}. Next, we will detail the global contrastive learning task and the local contrastive learning task.

\subsubsection{Global Contrastive Learning}
The global contrastive learning aims at helping node representation capture global information of the whole graph. Overall, the global contrastive learning is  based on maximizing the mutual information between the local node representation and the global graph representation. Specifically, as shown in Figure~\ref{fig:model-frame}, given an input graph sample $G$, different views can be generated via various types of data augmentations. For global contrastive learning, two views are adopted: one is the raw view $\text{View}_0$, where no changes are made on the original graph; the other one is augmented view $\text{View}_1$, where node attributes are randomly shuffled among all the nodes in a graph. With these two graph views, we can get two corresponding node representations ${\bf Z}_0$ and ${\bf Z}_1$ via a shared GNN model. After that, a global graph representation ${\bf g}_0$ can be summarized from the node representation matrix ${\bf Z}_0$ from the raw view $\text{View}_0$ via a multiplayer perceptron. 

Following~\cite{velickovic2019deep}, the positive samples in the global contrastive learning consist of node-graph representation pairs, where both the node representation and graph representation come from the raw view $\text{View}_0$. The negative samples consist of node-graph representation pairs where node representations come from $\text{View}_1$ and graph representation comes from $\text{View}_0$. A discriminator $\mathcal{D}$ is employed to compute the probability score for each pair, and the score should be higher for the positive pair and lower for the negative pair. In our work, we set  $\mathcal{D}({{\bf Z}_s}_i,{\bf g_0}) = Sigmoid({{\bf Z}_s}_i * {\bf g}_0)$, where ${\bf Z_s}_i$ denotes the representation of node $i$ from $\text{View}_s$ and $*$ denotes inner product. 

The objective function for the global contrastive learning is summarized as follows:

\begin{align}
    \mathcal{L}_g = -\frac{1}{2N}(\sum_{i=1}^{N}(log\mathcal{D}({{\bf Z}_0}_i,{\bf g}_0) +log(1-\mathcal{D}({{\bf Z}_1}_i,{\bf g}_0))),
    \label{eq:global_contrastive_loss} 
\end{align}
 where $N$ denotes the number of nodes in the input graph.
 
\subsubsection{Local Contrastive Learning}
In global contrastive learning, the model aims at capturing the global information of the whole graph into the node representations. In other words, the model attempts to learn an invariant graph representation from a global level and blend it into the node representations. However, graph data consists of nodes with various attributes and distinct structural roles. To fully exploit the structure information of a graph, in addition to the invariant graph representation from a global level, it is also important to learn invariant node representations from a local level. To achieve this, we propose local contrastive learning, which is based on distinguishing different nodes from different augmented views of a graph. 

As shown in Figure~\ref{fig:model-frame}, given an input graph $G$, we can generate two views of the graph via data augmentation. Since the local contrastive learning is to distinguish whether two nodes from different views are the same node of the input graph, the adopted data augmentation should not make large changes on the input graph. To meet this requirement, we adopt two adaptive graph data mechanisms -- adaptive edge dropping and adaptive node attribute masking. Specifically, the edges are dropped based on the edge importance in the graph. The more important an edge is, the less likely it would be dropped. Likewise, the node attributes are masked based on the importance of each dimension of node attributes. The more important the attribute dimension is, the less likely it would be masked out. Intuitively, in this work, the importance of both the edges and node attributes are computed based on the structural importance. Specifically, the importance score of an edge is computed as the average degree of its connected nodes and the importance score of an attribute is the average of the products of its norm over all nodes and the degree of the nodes in the whole graph. 

After the data augmentation, two views of an input graph are generated and serve as the input of the shared GNN model. The outputs are two node representation matrices ${\bf Z}_2$ and ${\bf Z}_3$ corresponding to two views. Note that the basic goal of local contrastive learning is to distinguish whether two nodes from augmented views are the same node in the input graph. Therefore, $({{\bf Z}_2}_i,{{\bf Z}_3}_i)$ ($i \in \{1,\dots,N\}$) denotes a positive pair, where $N$ is the number of nodes. $({{\bf Z}_2}_i,{{\bf Z}_3}_j)$ and $({{\bf Z}_2}_i,{{\bf Z}_2}_j)$ ($i,j \in \{1,\dots,N\}$ and $i\neq j$) denote an intra-view negative pair and an inter-view negative pair, respectively. Inspired by InfoNCE~\cite{oord2018representation,zhu2021graph}, the objective for a positive node pair $({{\bf Z}_2}_i,{{\bf Z}_3}_i)$ is defined as follow:
\begin{align}
    \mathfrak{l}_c\left({{\bf Z}_2}_i,{{\bf Z}_3}_i\right) = log\frac{h({{\bf Z}_2}_i,{{\bf Z}_3}_i)}{h({{\bf Z}_2}_i,{{\bf Z}_3}_i)+\sum_{j\neq i}h({{\bf Z}_2}_i,{{\bf Z}_3}_j)+\sum_{j\neq i}h({{\bf Z}_2}_i,{{\bf Z}_2}_j)},
\end{align}
where $h({{\bf Z}_2}_i,{{\bf Z}_3}_j) = e^{cos\left( g({{\bf Z}_2}_i),g({{\bf Z}_3}_j)\right)/ \tau}$, $cos()$ denotes the cosine similarity function, $\tau$ is a temperature parameter and $g()$ is a two-layer perceptron (MLP) to further enhance the model expression power. 
The node representations refined by this MLP are denoted as 
${\bf Z}_2^{'}$ and ${\bf Z}_3^{'}$, respectively. 

Apart from the contrastive objective described above, a decorrelation regularizer has also been added to the overall objective function of the local contrastive learning, in order to encourage different representation dimensions to capture distinct information. The decorrelation regularizer is applied over the refined node representations ${\bf Z}_2^{'}$ and ${\bf Z}_3^{'}$ as follows:
\begin{align}
    \mathfrak{l}_d({\bf Z}_2^{'}) =\left\| {{\bf Z}_2^{'}}^{T}{\bf Z}_2^{'} - I   \right\| ^{2}_F,
\end{align}
where $I$ denotes an identity matrix. The overall objective loss function of the local contrastive learning is denoted as follows:
\begin{align}
    \mathcal{L}_l = -
    \frac{1}{2N}\sum_{i=1}^N\left(\mathfrak{l}_c({{\bf Z}_2}_i,{{\bf Z}_3}_i) +\mathfrak{l}_c({{\bf Z}_3}_i,{{\bf Z}_2}_i) \right) + \frac{\beta}{2} (\mathfrak{l}_d({\bf Z}_2^{'})+\mathfrak{l}_d({\bf Z}_3^{'})),
    \label{eq:self-local}
\end{align}
where $\beta$ is the balance parameter. 

\subsubsection{The Overall Objective Function}
The overall objective loss function for the SSL task for GT3 is a weighted combination of the global contrastive learning loss and the local contrastive loss:
\begin{align}
    \mathcal{L}_s= \mathcal{L}_g +  \alpha \mathcal{L}_l,
    \label{eq:self-loss}
\end{align}
where $\alpha$ is the parameter balancing the local contrastive learning and the global contrastive learning. By minimizing the SSL loss defined in Eq~(\ref{eq:self-loss}), the model not only captures the global graph information but also learns the invariant and crucial node representations.
\subsection{Adaptation Constraint}
\begin{figure}[!t]
\begin{center}
{\includegraphics[width=0.95\linewidth]{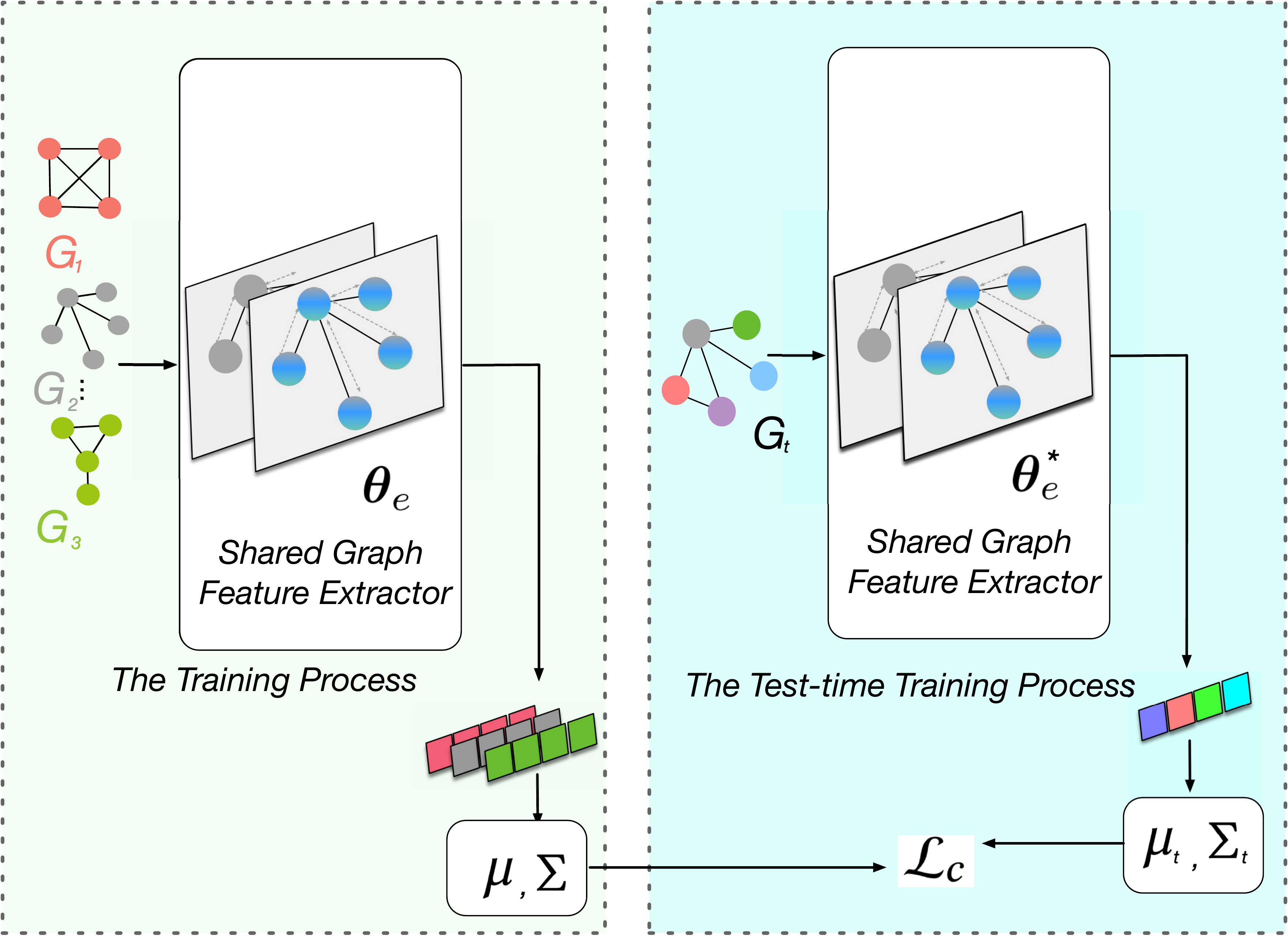}}
\end{center}

\caption{The Adaptation Constraint.}

\label{fig:model-cons}

\end{figure}

Directly applying test-time training could lead to severe representation distortion, since the model could overfit to the SSL task for a specific test sample~\cite{liu2021ttt++}, which hinders the model performance for the main task. This problem may be riskier  for test-time training for graph data since graph samples can vary from each other in not only node attributes but also graph structures. To mitigate this issue, we propose to add an adaptation constraint into the objective function during the test-time training process. The core idea is to put a constraint over the embedding spaces outputted by the \textit{shared graph feature extractor} between the training graph samples and the test graph sample, preventing the graph representations from severe fluctuations. 
In such a way, there is an additional constraint for the embedding of the test graph sample to be close to the embedding distribution of training graph samples which is generated with the guidance of the main task, in addition to just fitting the SSL task. 

The \textit{shared graph feature extractor} consists of the first $K$ GNN layers. Let us denote the node embeddings outputted by the \textit{shared graph feature extractor} for the training graphs $\{G_1,G_2,\ldots,G_n \}$ in the training process as $\{{\bf H}_1^K,{\bf H}_2^K,\ldots,{\bf H}_n^K\}$.  Once the training process completes, we first get graph embeddings via a read-out function ${\bf h}_{i}^K= READOUT({\bf H}_i^{K}) $, and then summarize two statistics of these graph embeddings: 
the empirical mean $\mu = \frac{1}{n}\sum_i^n {\bf h}_i^K$ and the covariance matrix ${\bm \Sigma} = \frac{1}{n-1}({{\bf H}^K}^T{{\bf H}^K}-({{\bf I}^T}{{\bf H}^K})^T{({{\bf I}^T}{{\bf H}^K})})$, where ${\bf H}^K = \{{{\bf h}_1^K}^T,\ldots,{{\bf h}_n^K}^T\}$. During the test-time training process, given an input test graph $G_t$, we also summarize embedding statistics of $G_t$ and its augmented views, and denote them as $\mu_t$ and ${\bm \Sigma}_t$. The adaptation constraint is to force the embedding statistics of the test graph sample to be close to these of training graph samples, which can be formally defined as: 
\begin{align}
    \mathcal{L}_c = \left\| \mu - \mu_t \right\|^2_2 + \left\| {\bm \Sigma} - {\bm \Sigma}_t \right\|^2_F.
\end{align}

\section{Theoretical Analysis}

In this section, we explore the rationality of test-time training for GNNs from a theoretical perspective. The goal is to demonstrate that the test-time training framework can be beneficial for GNNs in the graph classification task. Next, we first describe some basic notions and then develop two supportive theorems. 

For a graph classification task, suppose that there are $C$ classes and we use a $L$-layer simplified graph convolution (SGC)~\cite{wu2019simplifying} with the sum-pooling layer to learn graph representation, i.e., 
\begin{equation}
{\bf g}= f_\text{GNN}({\bf A}, {\bf X};{\bm \theta})= \text{pool}({\bf A}^L{\bf X}{\bf W}_1){\bf W}_2 = \bf{1}^T{\bf A}^L{\bf X}{\bm \theta},
\label{eq:SGC}
\end{equation}
where $\bf A$ and $\bf X$ are the input graph adjacency matrix and node attributes, $\bf{1}$ is a vector with all elements as $1$, and ${\bm \theta} ={\bf W}_1{\bf W}_2 $ denotes the model parameters. Note that we use SGC for this theoretical analysis since SGC has a similar filtering pattern as GCN but its architecture is simpler. Then, the $\textit{softmax}$ function is applied over the graph representation $ {\bf g}$ to get the prediction probability ${\bf p}_c$: 

\begin{equation}
{\bf p}_c = \frac{{e^{{\bf g}_c}}}{{\sum_{i=1}^C}e^{{\bf g}_i}},
\label{eq:softmax}
\end{equation}
where  $ \sum_{c=1}^C {\bf p}_c = 1 $. 
The cross-entropy loss is adopted as the optimization objective for the SGC model, i.e.,
\begin{equation}
\mathcal{L}({\bf A}, {\bf X},{y};{\bm  \theta})= -\sum_{c=1}^C 1_{y=c} \log({\bf p}_c),
\label{eq:cross_entro}
\end{equation}
where $y$ is the graph ground-truth label.

\begin{theorem}
Let $\mathcal{L}({\bf A}, {\bf X}, y;{\bm \theta})$ be defined based on Eqs~\ref{eq:SGC} and \ref{eq:softmax}, \ref{eq:cross_entro}, i.e., the cross-entropy loss for softmax classification for a $L$-layer SGC with sum-pooling for a graph ${\bf A}, {\bf X}, y $. It is convex and $\beta$-smooth in ${\bm \theta}$, and $\|\nabla_{\bm \theta}\mathcal{L}({\bf A}, {\bf X}, y;{\bm  \theta}) \| \leq G^{'}$ for all ${\bm \theta}$, where $G^{'}$ is a positive constant.
\label{the_1}
\end{theorem}

\textit{Proof Sketch.} We demonstrate that the cross-entropy loss for softmax classification is convex and $\beta$-smooth by proving its Hessian Matrix is positive semi-definite and the eigenvalues of the Hessian Metrix  are smaller than a positive constant. As $f_\text{GNN}$ defined in Eq~\ref{eq:SGC} is a linear transformation mapping function, thus it will not change these properties, which completes the proof of Theorem~\ref{the_1}. Proof details can be found in Appendix~\ref{proof_1}.

\begin{theorem}
Let $\mathcal{L}_m(x,y;{\bm \theta})$ denote a supervised task loss on a instance $x, y$ with parameters $\bm \theta$, and $\mathcal{L}_s(x;{\bm \theta})$ denotes a self-supervised task loss on a instance $x$ also with parameters $\bm \theta$. Assume that for all $x, y$, $\mathcal{L}_m(x,y;{\bm \theta})$ is differentiable, convex and $\beta$-smooth in ${\bm \theta}$, and both $\|\nabla_{{\bm \theta}}\mathcal{L}_m(x, y;{\bm  \theta}) \|, \|\nabla_{{\bm \theta}}\mathcal{L}_s(x;{\bm  \theta}) \| \leq G$ for all ${\bm \theta}$, where $G$ is a positive constant. With a fixed learning rate $\eta = \frac{\epsilon}{\beta G^2}$, for every $x, y$ such that
\begin{equation}
    	\left \langle \nabla_{{\bm \theta}}\mathcal{L}_m(x, y;{\bm  \theta}),\nabla_{{\bm \theta}}\mathcal{L}_s(x;{\bm  \theta})  \right \rangle >\epsilon,
\end{equation}
we have
\begin{equation}
    \mathcal{L}_m(x,y;{\bm \theta}) > \mathcal{L}_m(x,y;{\bm \theta}(x)),
\end{equation}
where ${\bm \theta}(x) = {\bm \theta} - \eta \nabla_{{\bm \theta}}\mathcal{L}_s(x;{\bm  \theta})$.
\label{the_2}
\end{theorem}

The detailed proof follows~\cite{sun2019test}, which is provided in Appendix~\ref{proof_2}. \textit{Theorem}~\ref{the_2} suggests that if the objective function of the main task is differentiable, convex, and $\beta$-smooth,  there exists an SSL task meeting some requirements and the test-time training via this SSL task can make the main task loss decrease. 

\textit{Remark.} From \textit{Theorem}~\ref{the_1}, we know that the objective loss function for the SGC model for graph classification is differentiable, convex and $\beta$-smooth. Combined with \textit{Theorem}~\ref{the_2}, we can conclude that test-time training can be beneficial for the GNN performance on graph classification. 
\section{Experiment}

In this section, we validate the effectiveness and explore the rationality of the proposed GT3 via empirical studies. In the following subsections, the experimental settings is first introduced. Then, we conduct a layer-wise feature space exploration of GNN models trained for different tasks to further validate the rationality of GT3. Next, the classification performance comparison of the proposed GT3 and baseline methods is discussed. Finally, we perform an ablation study and hyper-parameter sensitivity exploration.

\subsection{Experimental Settings}
In this study, we implement the proposed GT3 on two well-known GNN models: the Graph Convolutional Network (GCN)~\cite{kipf2016semi} and the Graph Isomorphism Network (GIN)~\cite{xu2018how}. Empirically, we implement four-layer GCN models and three-layer GIN models with max-pooling. The layer normalization is adopted in the implementations. The hyper-parameters are selected based on the validation set from the ranges listed as follows:
\begin{itemize}
    \item Hidden Embedding Dimension : {32,64,128}
    \item Batch Size: {8,16,32,64}
    \item Learning Rate: [1e-5, 5e-3]
    \item Dropout Rate: {0.0,0.1,0.2,0.3,0.4,0.5}
\end{itemize}

For the balance hyper-parameters $\alpha$, $\beta$ and $\gamma$ in the SSL task, we determine their values based on the joint training of the SSL task and the main task on the validation set. As for performance evaluation, we conduct experiments on data splits generated by 3 random seeds and  model initialization of 2 random seeds on DD, ENZYMES, and PROTEINS, and report the average and variance values of classification accuracy. For ogbg-molhiv, we do an empirical study on the fixed official data split with  model initialization of 2 random seeds and report the average performance in terms of ROC-AUC. We conduct experiments on four popular graph datasets: 
\begin{itemize}
    \item \textbf{DD}~\cite{morris2020tudataset}: a dataset consists of protein structures of two categories, where each node denotes an amino aid.
    \item \textbf{ENZYMES}~\cite{morris2020tudataset}: it consists of enzymes of six categories, where each enzyme is represented as its tertiary structure. 
    \item \textbf{PROTEINS}~\cite{morris2020tudataset}: a dataset contains protein structures of two categories, where each node denotes a secondary  structure  element.
    \item \textbf{ogbg-molhiv}~\cite{hu2020open}: it includes molecules of two categories (carrying HIV virus or not), where each node is an atom.
\end{itemize}

In order to simulate the scenario where the test set distribution is different from the training set, graph samples from {DD}, {ENZYMES}, and {PROTEINS} are split into two groups based on their graph size (number of nodes). Next, we randomly select 80\% of the graphs from the small group as the training set and the other 20\% of the graphs as the validation set. The test set consists of graphs randomly chosen from the larger group and the number of test graphs is the same as that of the validation set. For the {ogbg-molhiv}, we adopt the official scaffold splitting, which also aims at separating graphs with different structures into different sets. For simplicity, these data splits described above are denoted as \textit{OOD} data split. To validate the reasonability of the \textit{OOD} data split, we examine the training, validation, and test performance of GCN and the results are shown in Table \ref{tab:performance_gap}. Note that the performance in ogbg-molhiv is copied from ~\cite{hu2020open}. 
One can clearly see that there exists a significant performance gap between the training set and the test set, verifying the motivation of this work.   

\begin{table}[t]

\centering
\caption{Graph Classification Performance Gap of GCN on Training Set, Validation Set, and Test Set based on  \textit{OOD} Data Split. Note that the performance metric for ogbg-molhiv is \textbf{ROC-AUC}(\%) and that for others is \textbf{Accuracy}(\%).}
\begin{adjustbox}{width = 0.45\textwidth}

  \setlength{\tabcolsep}{4mm}{
\begin{tabular}{cccc}

\hline
\multirow{2}{*}{\textbf{Dataset}} & \multicolumn{3}{c}{\textbf{Graph Classification Performance}} \\ 
     & Training & Validation  & Test    \\ \hline
DD & 92.8\scriptsize$\pm{7.13}$ & 73.82 \scriptsize$\pm{3.36}$          & 49.43 \scriptsize$\pm{15.30}$   \\ \hline
ENZYMES  & 87.43\scriptsize$\pm{12.34}$    &65.22 \scriptsize$\pm{3.76}$          & 37.43 \scriptsize$\pm{5.03}$   \\ \hline
PROTEINS  & 77.5\scriptsize$\pm{2.07}$    & 77.78 \scriptsize$\pm{1.93}$          & 68.46 \scriptsize$\pm{7.17}$   \\ \hline
ogbg-molhiv  & 88.65\scriptsize$\pm{1.01}$    & 83.73\scriptsize$\pm{0.78}$          & 74.18\scriptsize$\pm{1.22}$   \\ \hline

\end{tabular}
}
 \end{adjustbox}
\label{tab:performance_gap}

\end{table}

\begin{figure}[htb]%

    \centering
    \subfigure[DD\label{fig:DD-main}]{{\includegraphics[width=0.55\linewidth, trim = 0.8cm 0.5cm 1cm 1cm, clip]{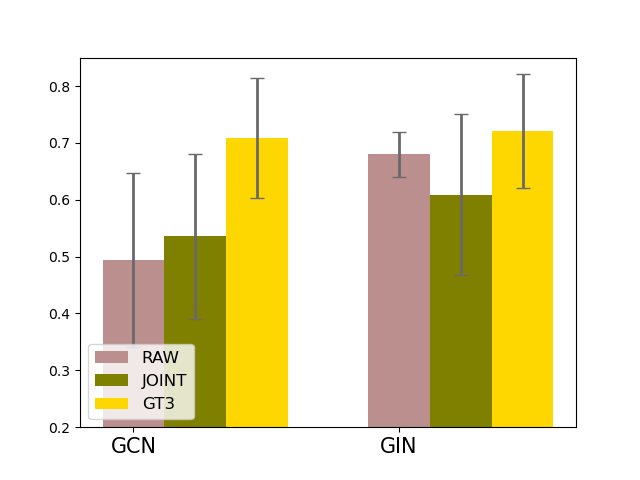}}}%
    \subfigure[ENZYMES\label{fig:ENZ-main}]{{\includegraphics[width=0.55\linewidth, trim = 0.8cm 0.5cm 1cm 1cm, clip]{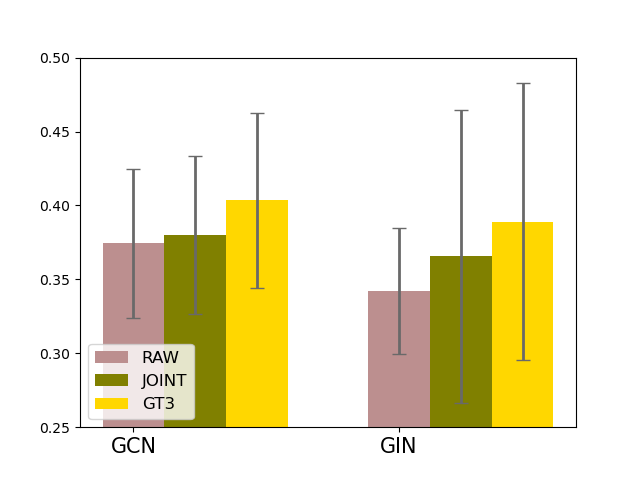} }}%

    \subfigure[PROTEINS \label{fig:PRO-main}]{{\includegraphics[width=0.55\linewidth, trim = 0.8cm 0.5cm 1cm 1cm, clip]{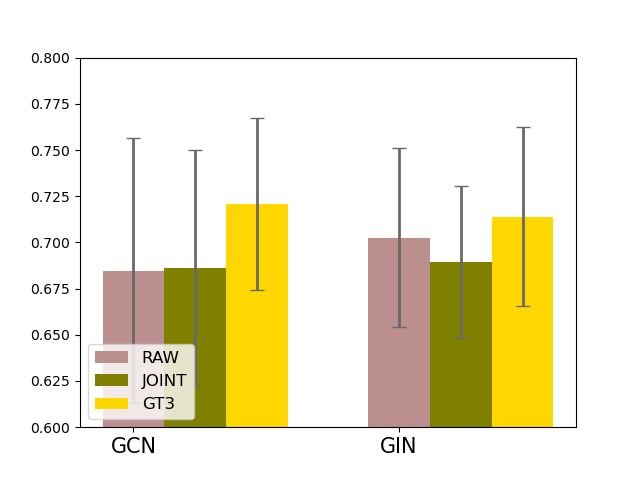}}}%
    \subfigure[ogbg-molhiv \label{fig:ogbg-main}]{{\includegraphics[width=0.55\linewidth, trim = 0.8cm 0.5cm 1cm 1cm, clip]{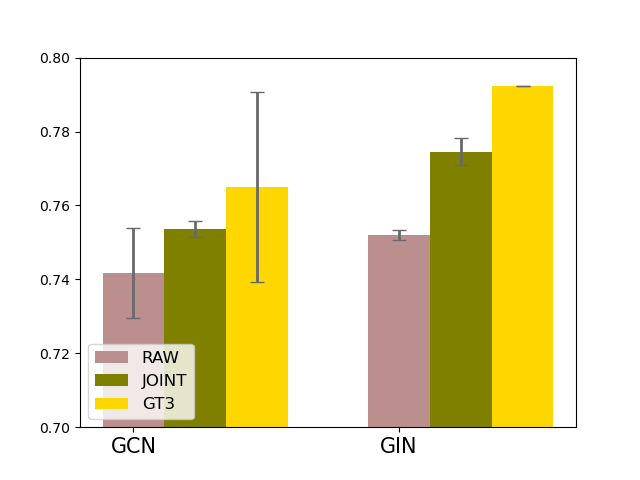} }}%
    \qquad

    \caption{Performance Comparison of GCN and GIN on Four Datasets based on \textit{OOD} Data Split. Note that the performance metric for ogbg-molhiv is \textbf{ROC-AUC}(\%) and that for others is \textbf{Accuracy}(\%).}%

    \label{fig:main-results}

\end{figure}

\subsection{Performance Comparison}

In this subsection, we compare the graph classification performance of GCN and GIN in three mechanisms: (1) \textit{RAW} model is trained only by the main task, (2) \textit{JOINT} model is jointly trained by the main task and the proposed two-level SSL task, and (3) \textit{GT3} model is learned by test-time training. The comparison results are shown in Figure~\ref{fig:main-results}. Note that the performance of \textit{RAW} in ogbg-molhiv is copied from~\cite{hu2020open}. From the figure, we can observe that the joint training of the proposed two-level SSL task and the main task can slightly improve the model performance in some cases. However, the proposed \textit{GT3} is able to consistently perform better than both \textit{RAW} and \textit{JOINT}, which demonstrates the effectiveness of the proposed \textit{GT3}. To further demonstrate the performance improvement from \textit{GT3}, we summarize the relative performance improvement of \textit{GT3} over \textit{RAW}. The results are shown in Table~\ref{tab:performance_improvement} where the performance improvement can be up to $43.3\%$. In addition to the performance comparison on the \textit{OOD} data split, we have also conducted a  performance comparison on a random data split, where we randomly split the dataset into 80\%/10\%/10\% for the training/validation/test sets, respectively. The results demonstrate that GT3 can also improve the classification performance in most cases, and more results can be found in Appendix~\ref{app:std}.

\begin{table}[b]

\centering
\caption{Relative Performance Improvement of \textit{GT3} over \textit{RAW}.}
\begin{adjustbox}{width = 0.45\textwidth}

  \setlength{\tabcolsep}{4mm}{
\begin{tabular}{ccccc}

\hline
\multirow{2}{*}{\textbf{Models}} & \multicolumn{4}{c}{\textbf{Datasets}} \\ 
     & DD & ENZYMES & PROTEINS & ogbg-molhiv    \\ \hline
GCN & +43.3\% & +7.81\% & +5.25\% & +3.13\% \\ \hline
GIN  & +6.09\% & +13.67\% & +1.63\% & +5.21\%   \\ \hline

\end{tabular}
}
 \end{adjustbox}
\label{tab:performance_improvement}

\end{table} 

\subsection{Layer-wise Representation Exploration}\label{exp:layer-fea-explore}
Test-time training framework is designed to share the first few layers of Convolutional Neural Networks (CNNs) in the existent work. This design is very natural since it is well received that the first few layers of CNNs capture similar basic patterns even when they are trained for different tasks. However, it is unclear how GNNs work among different tasks. To answer this question, we explore the representation similarity of GNNs trained for different tasks. To be specific, we train four GCN models with exactly the same architectures on ENZYMES and PROTEIN for four different tasks: the graph classification task defined in Eq~\ref{eq:main_task}, the proposed GT3 SSL task with loss described in Eq~(\ref{eq:self-loss}), the SSL tasks with the local contrastive loss (Eq~\ref{eq:self-local}) and the global contrastive loss (Eq~\ref{eq:global_contrastive_loss}). Then we leverage Centered Kernel Alignment (CKA)~\cite{kornblith2019similarity} as a similarity index to evaluate the representation similarity of different GCN models layer by layer. 

\begin{figure}[ht]%

    \centering

    \subfigure[ENZYMES. \label{fig:ENZYMES-CKA}]{{\includegraphics[width=0.55\linewidth, trim = 2cm 0.5cm 1cm 1.5cm, clip]{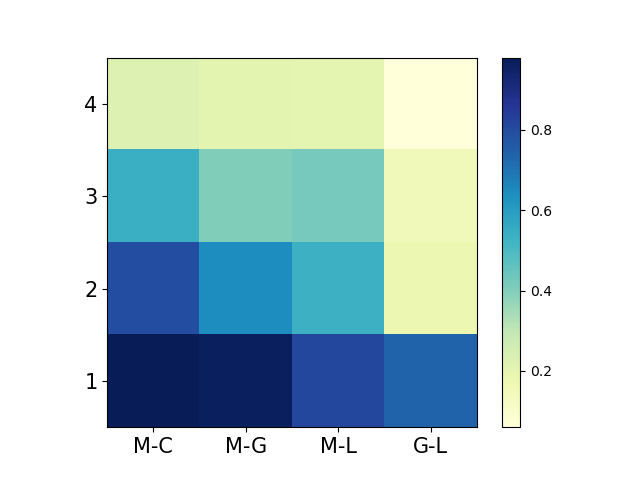}}}%
    \subfigure[PROTEINS. \label{fig:PRO_CKA}]{{\includegraphics[width=0.55\linewidth, trim = 2cm 0.5cm 1cm 1.5cm, clip]{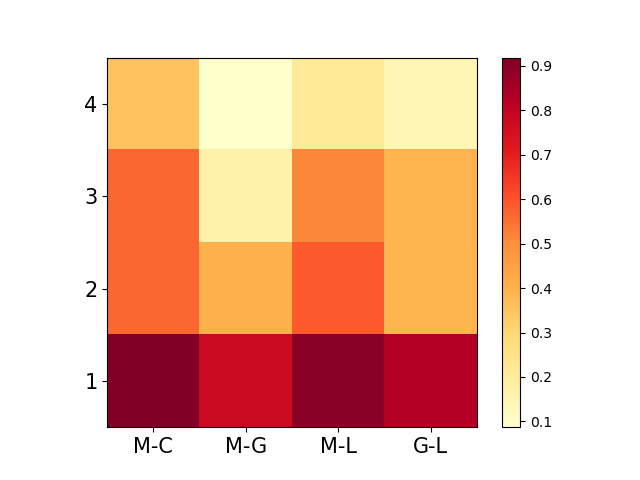} }}%

    \caption{CKA Values of Representation from Different Layers for Various Tasks. Note that "M" denotes the graph classification task; "C" indicates the proposed GT3 SSL task; "G" is the global contrastive learning task; and "L" is the local contrastive learning task.}%
    \label{fig:cka}

\end{figure}

% \jt{I think we need to explain what M-C... mean in the figure.}
The similarity results are demonstrated in Figure~\ref{fig:cka}, where ``M" denotes the graph classification task,  ``C" indicates the proposed GT3 SSL task, ``G" is the global contrastive learning task and ``L" is the local contrastive learning task. "M-C" represents the representation pair of "M" and "C" to be compared. The representation pair for other tasks are denoted similarly. We can make the following observations. First, the representations for different tasks tend to be more similar to each other in the lower layer. Second, the representation similarity for the same task pair may be different in different datasets. The representations of the graph classification task and the global contrastive learning task are more similar to each other compared to that of the graph classification task and the local contrastive learning task in ENZYMES, while the observation is the opposite in PROTEINS. Third, the representation similarity between the local contrastive learning and the global contrastive learning is relatively lower, which reveals that they may learn knowledge that is  complementary to each other. These observations motivate us (1) to share the first few layers of GNNs in test-time training; and (2) to develop the two-level SSL task for GT3, which is capable of capturing graph information from both the local and global levels.

\subsection{Ablation Study}

\begin{figure}[ht]%

    \centering

    \subfigure[The Ablation Study of GT3. \label{fig:ablation}]{{\includegraphics[width=0.55\linewidth, trim = 2cm 0.5cm 1cm 1.5cm, clip]{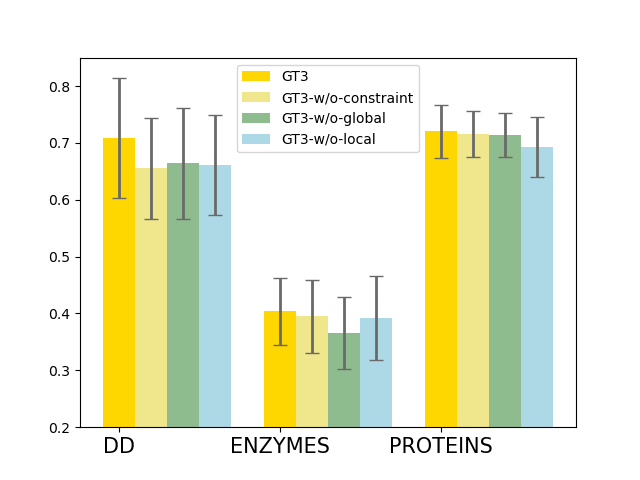}}}%
    \subfigure[The Effect of Different Shared Layers. \label{fig:number-layer}]{{\includegraphics[width=0.5\linewidth, trim = 0.5cm 0cm 0cm 0cm, clip]{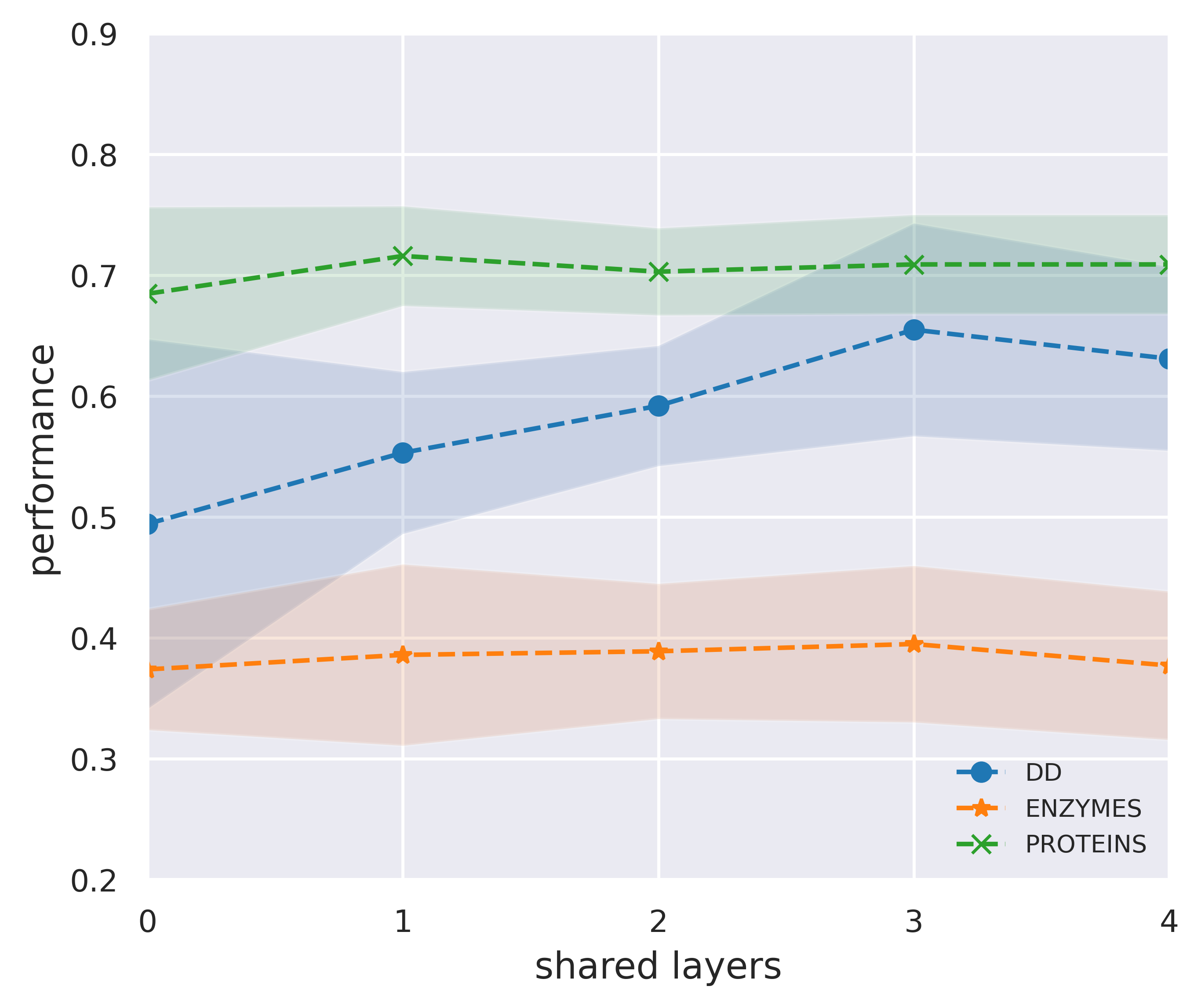} }}%
    \caption{(a)The Ablation Study of GT3 and (b) The Effect of Different Shared Layers.}

\end{figure}

In this subsection, we conduct an ablation study to understand the impact of key components on the performance of the proposed GT3 including the SSL task and the adaptation constraint. This ablation study is conducted on DD, ENZYMES, and PROTEINS with the GCN model. Specifically, we study the effect of the adaptation constraint, global contrastive learning, and local contrastive learning by respectively eliminating each of them from GT3. These three variants of GT3 are denoted as GT3-w/o-constraint, GT3-w/o-global, and GT3-w/o-local. The performance is shown in Figure~\ref{fig:ablation}. We observe that the removal of each component will degrade the model's performance. This demonstrates the necessity of all three components.

\subsection{The Effect of Shared Layers}
We have also explored the effect of the number of shared layers on the performance of GT3. To be specific, we vary the number of shared layers of the GCN model in GT3 as $\{0,1,2,4\}$. Note that the model where the number of shared layers is zero denotes the model trained only by the main task. The graph classification performance on DD, ENZYMES, and PROTEINS is shown in Figure~\ref{fig:number-layer}. From the figure, we can observe that most of the GT3 implementations with different shared layers can improve the classification performance over their baseline counterparts, while an appropriate choice of shared layers can boost the model performance significantly on some datasets such as DD.

\section{Related Work}
\textbf{Test-Time Training.}
Test-time training with self-supervision is proposed in~\cite{sun2019test}, which aims at improving the model generalization under distribution shift via solving a self-supervised task for test samples. This framework has empirically demonstrated its effectiveness in bridging the performance gap between training set and test set in image domain~\cite{sun2019test,bartler2021mt3,fu2021learning}. There are also works combining the test-time training and meta-learning learning~\cite{bartler2021mt3}, and  attempting to apply test-time training in reinforcement learning~\cite{hansen2020self}. Apart from these applied researches, a recent work tries to explore when test-time training thrives from a theoretical perspective~\cite{liu2021ttt++}.  

\textbf{Graph Neural Networks.}
Graph Neural Network is a successful generalization of Deep Neural Network over graph data. It has been empirically and theoretically proven to be very powerful in graph representation learning~\cite{xu2018how}, and has been widely used in various applications, such as recommendation~\cite{wang2021localized,he2020lightgcn,fan2019graph}, social network analysis~\cite{tan2019deep,huang2022going}, natural language processing~\cite{huang2019text,cai2020graph,li2022house} and sponsored search~\cite{pang2022improving,li2021adsgnn}. GNNs typically refine node representations by feature aggregation and transformation. Most existent GNN models can be summarized into a message-passing framework consisting of message passing and feature update~\cite{gilmer2017neural}. There is also another perspective unifying many GNN models as methods to solve a graph signal denoising problem~\cite{ma2020unified}. Overall, GNN is active research filed and there are a lot of works focusing on improving the performance, robustness, and scalability of the GNN models~\cite{chen2020scalable,jin2020graph,liu2021elastic}. 
\section{Conclusion}
In this work, we design a test-time training framework for GNNs (GT3) for graph classification task, with the aim to bridge the performance gap between the training set and the test set when there is a data distribution shift. As the first work to combine test-time training and GNNs, we empirically explore the rationality of the test-time training framework over GNNs via a layer-wise representation comparison of GNNs for different tasks. In addition, we validate that the GNN performance on graph classification can be benefited by test-time training from a theoretical perspective. Furthermore, we demonstrate the effectiveness of the proposed GT3 on extensive experiments and understand the impact of the model components via ablation study. As the future work, it is worth studying  test-time training for GNNs for other tasks on graphs such as node classification. Also, it is interesting to explore how to automatically determine the number of shared layers and choose appropriate SSL tasks for various main tasks on different datasets.

\bibliographystyle{ACM-Reference-Format}
\bibliography{sample-base}

\newpage

\appendix
\section{Appendix}
\subsection{Theorem Proof}
\subsubsection{Proof of Theorem 1}
\label{proof_1}
Denote $\mathcal{L}(y;{\bf g}) = -\sum_{c=1}^C 1_{y=c} \log(\frac{{e^{{\bf g}_c}}}{{\sum_{i=1}^C}e^{{\bf g}_i}})$, and from Eq~\ref{eq:SGC}, we have 
\begin{equation}
{\bf g}= f_\text{GNN}({\bf A}, {\bf X};{\bm \theta})= \bf{1}^T{\bf A}^L{\bf X}{\bm \theta}. 
\label{eq:z_over_theta}
\end{equation}
\begin{align}
\mathcal{L}(y;{\bf g}) & =   -\sum_{c=1}^C 1_{y=c} \log(\frac{{e^{{\bf g}_c}}}{{\sum_{i=1}^C}e^{{\bf g}_i}}) \nonumber \\ 
& = -\log(\frac{e^{\bf{z}_c}}{\sum_{i=1}^{C}{e^{\bf{z}_i}}}).
\label{eq:L_over_z}
\end{align}
The Hessian Matrix of Eq~\ref{eq:L_over_z}, we have
\begin{equation}
    \mathcal{\bf H}_{ij}=
   \begin{cases}
   -{\bf p}_i{\bf p}_j& \text{i}\neq\text{j}\\
   {\bf p}_i-{\bf p}_i^2& \text{i=j},
\end{cases}
\label{eq:hessian}
\end{equation}
where ${\bf p}_i$ is defined in Eq~\ref{eq:softmax}. 

For $ \forall {\bf a} \in \mathbb{R}^C$, we have
\begin{align}
    {\bf a}^T\mathcal{\bf H}{\bf a} &= \sum_{i=1}^C{\bf p}_i{\bf a}_i^2 - (\sum_{i=1}^C{{\bf a}_i}{\bf p}_i)^2 \nonumber \\
    & =  \sum_{i=1}^C{\bf p}_i{\bf a}_i^2 - (\sum_{i=1}^{C}{\bf a}_i\sqrt{{\bf p}_i}\sqrt{{\bf p}_i})^2
\end{align}

According to Cauchy-Schwarz Inequality, we have
\begin{align}
(\sum_{i=1}^C{\bf a}_i\sqrt{{\bf p}_i}\sqrt{{\bf p}_i})^2 \leq (\sum_{i=1}^C{{\bf a}_i}^2{\bf p}_i)(\sum_{i=1}^C{\bf p}_i) .
\end{align}

Thus, we have 
\begin{align}
    {\bf a}^T\mathcal{\bf H}{\bf a} &\geq \sum_{i=1}^C{\bf p}_i{\bf a}_i^2 - (\sum_{i=1}^{C}{{\bf a}_i}^2{\bf p}_i)(\sum_{i=1}^C {\bf p}_i) .
\end{align}

According to the definition of ${\bf p}_i$ in Eq~\ref{eq:softmax}, we have 
\begin{align}
\sum_{i=1}^C{\bf p}_i = 1 . 
\end{align}

Therefore,
\begin{align}
    {\bf a}^T\mathcal{\bf H}{\bf a} &\geq 0 .
\end{align}

Thus, we have proved that the Hessian Matrix of the function defined in Eq~\ref{eq:L_over_z}, i.e., $\mathcal{\bf H}$, is positive semi-definite (PSD). This is equivalent to prove that $\mathcal{L}(y;{\bf g})$ is \textbf{convex} in $\bf g$ (\textbf{property 1}).

For $ \forall {\bf b} \in \mathbb{R}^C$ and $ \|{\bf b}\| = 1$,   we have
\begin{align}
    {\bf b}^T\mathcal{\bf H}{\bf b} &= \sum_{i=1}^C{\bf p}_i{\bf b}_i^2 - (\sum_{i=1}^C{{\bf b}_i}{\bf p}_i)^2 \nonumber \\
    & =  \sum_{i=1}^C{\bf p}_i{\bf b}_i^2 - (\sum_{i=1}^{C}{\bf b}_i\sqrt{{\bf p}_i}\sqrt{{\bf p}_i})^2 \nonumber \\
    & \leq \sum_{i=1}^C{\bf b}_i^2 + (\sum_{i=1}^{C}{\bf b}_i\sqrt{{\bf p}_i}\sqrt{{\bf p}_i})^2. 
\end{align}

According to Cauchy-Schwarz Inequality, we can arrive at 
\begin{align}
    {\bf b}^T\mathcal{\bf H}{\bf b} 
    & \leq \sum_{i=1}^C{\bf b}_i^2 + (\sum_{i=1}^{C}{\bf b}_i\sqrt{{\bf p}_i}\sqrt{{\bf p}_i})^2 \nonumber \\
    & \leq  \sum_{i=1}^C{\bf b}_i^2 + (\sum_{i=1}^{C}{\bf b}_i^2)(\sum_{i=1}^{C}{\bf p}_i).
\end{align}

Since $ \|{\bf b}\| = 1$ and $\sum_{i=1}^C{\bf p}_i = 1$, we have 
\begin{align}
    {\bf b}^T\mathcal{\bf H}{\bf b} 
    & \leq 2.
\end{align}

We have thus proved that the eigenvalues of the Hessian Matrix $\mathcal{\bf H}$ is smaller than 2. This is equivalent to prove that $\mathcal{L}(y;{\bf g})$ is \textbf{$\beta$-smooth} in $\bf g$ (\textbf{property 2}).

By computing the first-order gradient of $\mathcal{L}(y;{\bf g})$ over $\bf{z}$, we have
\begin{equation}
\nabla_{\bf g}{\mathcal{L}({\bf A}, {\bf X}, {y};{\bf g}})=
   \begin{cases}
   -1 + {\bf p}_i& \text{i=c}\\
   {\bf p}_i& \text{i} \neq\text{c}.
\end{cases}
\label{eq:first-order-gra} 
\end{equation}
Then, we can derive its $L_2$ norm
\begin{align}
    \| \nabla_{\bf g}{\mathcal{L}({y};{\bf g}}) \| &= \sqrt{(-1+{\bf p}_c)^2+ \sum_{i\neq c}({\bf p}_i)^2} \nonumber \\
    &= \sqrt{\sum_{i\neq c}({\bf p}_i)^2 + \sum_{i\neq c}({\bf p}_i)^2} \nonumber \\
    &\leq 2.
\end{align}

We have thus proved that the $L_2$ norm of the first-order gradient of $\mathcal{L}(y;{\bf g})$ over $\bf{z}$ is bounded by the a positive constant (\textbf{property 3}).

So far, we have proved that for all ${\bf g}, y$, $\mathcal{L}(y;{\bf g})$ is convex and $\beta$-smooth in ${\bf g}$, and both $\|\nabla_{\bf g}\mathcal{L}(y;{\bf g}) \| \leq G$ for all ${\bf g}$, where $G$ is a positive constant. According to Eq~\ref{eq:z_over_theta}, $f_\text{GNN}$ is a linear transformation mapping function and won't change these three properties. Therefore, the proof of Theorem 1 is completed.

\begin{figure*}[htb]%

    \centering
    \subfigure[DD\label{fig:DD-main}]{{\includegraphics[width=0.31\linewidth]{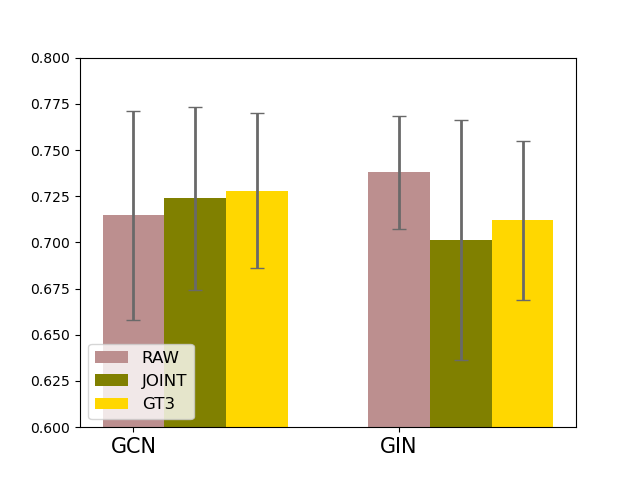}}}%
    \subfigure[ENZYMES\label{fig:ENZ-main}]{{\includegraphics[width=0.31\linewidth]{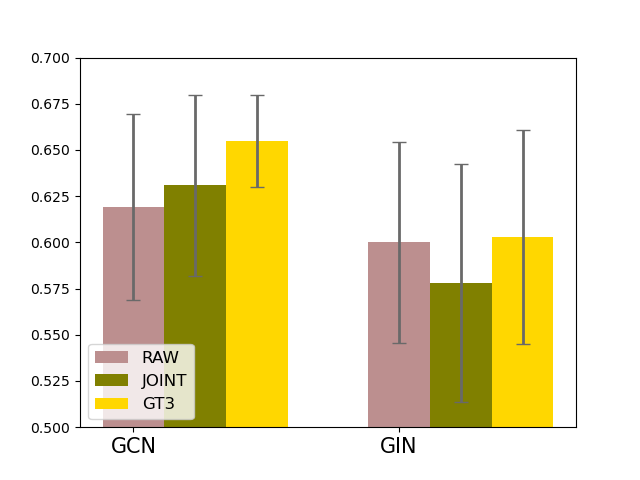} }}%
    \subfigure[PROTEINS \label{fig:PRO-main}]{{\includegraphics[width=0.31\linewidth]{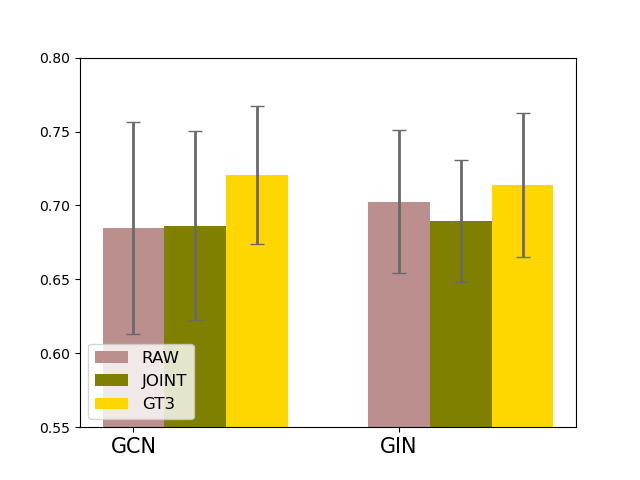}}}%
    \caption{Performance Comparison of GCN and GIN on Three Datasets on the Random Data Split. Note that the performance metric for ogbg-molhiv is \textbf{ROC-AUC}(\%) and that for others is \textbf{Accuracy}(\%).}%

    \label{fig:std-results}

\end{figure*}

\subsubsection{Proof of Theorem 2}
\label{proof_2}
We follow~\cite{sun2019test} to prove Theorem 2. For any $\eta$, by $\beta$-smoothness, we have 
\begin{flalign}
    &\mathcal{L}_m(x,y;{\bm \theta(x)}) = \mathcal{L}_m(x,y;{\bm \theta(x)} - \eta\nabla_{\bm \theta}{\mathcal{L}_s(x;{\bm \theta})}) \nonumber \\
    & \leq \mathcal{L}_m(x,y;{\bm \theta(x)} - \eta\left \langle \nabla_{{\bm \theta}}\mathcal{L}_m(x, y;{\bm  \theta}),\nabla_{{\bm \theta}}\mathcal{L}_s(x;{\bm  \theta})  \right \rangle \nonumber\\ 
    &+ \frac{\eta^2\beta}{2} \|\nabla_{\bm \theta}{\mathcal{L}_s(x;{\bm \theta})}\|^2.
\end{flalign}
Denote 
\begin{equation}
    \eta^* = \frac{\left \langle \nabla_{{\bm \theta}}\mathcal{L}_m(x, y;{\bm  \theta}),\nabla_{{\bm \theta}}\mathcal{L}_s(x;{\bm  \theta})  \right \rangle }{\beta  \|\nabla_{\bm \theta}{\mathcal{L}_s(x;{\bm \theta})}\|^2},
\end{equation}
then we have
\begin{flalign}
&\mathcal{L}_m(x,y;{\bm \theta} - \eta^{*}\nabla_{\bm \theta}{\mathcal{L}_s(x;{\bm \theta})}) \nonumber \\ 
&\leq \mathcal{L}_m(x,y;{\bm \theta})  - \frac{{\left \langle \nabla_{{\bm \theta}}\mathcal{L}_m(x, y;{\bm  \theta}),\nabla_{{\bm \theta}}\mathcal{L}_s(x;{\bm  \theta})  \right \rangle }^2}{2\beta \|\nabla_{\bm \theta}{\mathcal{L}_s(x;{\bm \theta})}\|^2)}.
\end{flalign}

By the assumptions on the $L_2$ norm of the first-order gradients for the main task and the ssl task and the assumption on their inner product, we have 

\begin{align}
    \mathcal{L}_m(x,y;{\bm \theta} ) -\mathcal{L}_m(x,y;{\bm \theta} - \eta^{*}\nabla_{\bm \theta}{\mathcal{L}_s(x;{\bm \theta})}) \geq \frac{\epsilon^2}{2\beta G^2}.
\end{align}

Denote $\eta = \frac{\epsilon}{\beta G^2} $, based on the assumptions, we have $0<\eta\leq\eta^*$. Denote ${\bf t}= \nabla_{\bm \theta}{\mathcal{L}_s(x;{\bm \theta})},$ by convexity of $\mathcal{L}_m$,
\begin{flalign}
&\mathcal{L}_m(x,y;{\bm \theta}(x)) = \mathcal{L}_m(x,y;{\bm \theta} - \eta{\bf t}) \nonumber \\
&= \mathcal{L}_m(x,y;(1-\frac{\eta}{\eta^*}){\bm \theta}+ \frac{\eta}{\eta^*}({\bm \theta}- \eta^*{\bf t})) \nonumber \\
&\leq (1-\frac{\eta}{\eta^*})\mathcal{L}_m(x,y;{\bm \theta}) + \frac{\eta}{\eta^*}\mathcal{L}_m(x,y;{\bm \theta} - \eta^*{\bf t}) \nonumber \\
&\leq (1-\frac{\eta}{\eta^*})\mathcal{L}_m(x,y;{\bm \theta}) + \frac{\eta}{\eta^*}(\mathcal{L}_m(x,y;{\bm \theta}) - \frac{\epsilon^2}{2\beta G^2})\nonumber \\
&= \mathcal{L}_m(x,y;{\bm \theta}) - \frac{\eta}{\eta^*}\frac{\epsilon^2}{2\beta G^2}.
\end{flalign}

Since $\frac{\eta}{\eta^*} > 0$, we have 
\begin{align}
    \mathcal{L}_m(x,y;{\bm \theta})-\mathcal{L}_m(x,y;{\bm \theta}(x)) > 0.
\end{align}

\subsection{Performance Comparison on Random Data Split}
\label{app:std}
 In addition to the performance comparison on the \textit{OOD} data split, we have also conducted performance comparison on a random data split, where we randomly split the dataset into 80\%/10\%/10\% for the training/validation/test sets, respectively. The results are shown in Figure~\ref{fig:std-results}. From it, we can observe that GT3 can also improve the classification performance in most cases, which demonstrates the effectiveness of the proposed GT3 in multiple scenarios.

\end{document}